\documentclass[bimj,fleqn]{w-art}
\usepackage{times}
\usepackage{w-thm}
\usepackage[]{algorithm, algorithmic}
\usepackage{graphicx}
\usepackage{amsmath, amssymb}
\usepackage{booktabs}
\usepackage{color}
\usepackage[usenames,dvipsnames]{xcolor}
\usepackage[round]{natbib}
\newlength\myindent 
\newcommand\bindent{%
  \begingroup 
  \setlength{\itemindent}{\myindent} 
  \addtolength{\algorithmicindent}{\myindent} 
}
\newcommand\eindent{\endgroup} %

\theoremstyle{plain}

\theoremstyle{definition}

\usepackage[]{graphicx}
\chardef\bslash=`\\ 

\begin{document}
\DOIsuffix{bimj.200100000}
\Volume{X}
\Issue{XX}
\Year{XXXX}
\pagespan{1}{}

\title[Boosting Joint Models]{Boosting Joint Models for Longitudinal and Time-to-Event Data}
\author[Elisabeth Waldmann {\it{et al.}}]{Elisabeth Waldmann\footnote{Corresponding author: {\sf{e-mail: elisabeth.waldmann@fau.de}}, Phone: +49-9131-85-22722, Fax: +49-9131-85-25740}\inst{,1}}
\address[\inst{1}]{Department of Medical Informatics, Biometry and Epidemiology,  Waldstra\ss e 6, 91054 Erlangen, Germany}
\author[]{David Taylor-Robinson\inst{2}}
\address[\inst{2}]{Department of Public Health and Policy, Farr Institute, University of Liverpool Waterhouse Building, Block B, Brownlow Street, Liverpool L69 3GL, United Kingdom}
\author[]{Nadja Klein \inst{3} }
\address[\inst{3}]{Chairs of Statistics and Econometrics, Georg-August-Universit\"at G\"ottingen, Humboldtallee 3,
37073 G\"ottingen, Germany}
\author[]{Thomas Kneib \inst{3} }
\author[]{Tania Pressler\inst{4}}
\address[\inst{4}]{Cystic Fibrosis Center, Rigshospitalet, Copenhagen, Denmark}
\author[]{Matthias Schmid \inst{5} }
\address[\inst{5}]{Department of Medical Biometrics, Informatics and Epidemiology, Rheinische Friedrich-Wilhelms-Universit\"at Bonn, Sigmund-Freud-Stra\ss e 25,
53105 Bonn, Germany}
\author[]{Andreas Mayr \inst{1,5} }
\Receiveddate{zzz} \Reviseddate{zzz} \Accepteddate{zzz}

\begin{abstract}

Joint Models for longitudinal and time-to-event data have gained a lot of attention in the last few years as they are a helpful technique to approach common a data structure in clinical studies where longitudinal outcomes are recorded alongside event times. Those two processes are often linked and the two outcomes should thus be modeled jointly in order to prevent the potential bias introduced by independent modelling. Commonly, joint models are estimated in likelihood based expectation maximization or Bayesian approaches using frameworks where variable selection is problematic and which do not immediately work for high-dimensional data. In this paper, we propose a boosting algorithm tackling these challenges by being able to simultaneously estimate predictors for joint models and automatically select the most influential variables even in high-dimensional data situations. We analyse the performance of the new algorithm in a simulation study and apply it to the Danish cystic fibrosis registry which collects longitudinal lung function data on patients with cystic fibrosis together with data regarding the onset of pulmonary infections. This is the first approach to combine state-of-the art algorithms from the field of machine-learning with the model class of joint models, providing a fully data-driven mechanism to select variables and predictor effects in a unified framework of boosting joint models.
\end{abstract}

\maketitle                   

\section{Introduction}

The terms \textit{joint models} or \textit{joint modelling} have been used in various contexts to describe modelling of a combination of different outcomes. This article deals with joint models for longitudinal and survival outcomes, in which the predictors for both  are composed of individual as well as {\textit{shared sub-predictors}} (i.e. a part of the predictor which is used in both, the longitudinal and the survival part of the model) . The shared sub-predictor is scaled by an {\textit{association parameter}} which quantifies the relationship between the two parts of the model. This type of model was first suggested by \cite{wulfsohn} in order to prevent the bias resulting from the independent estimation of the two entities, and this approach has been modified and extended subsequently in various ways. Simulation studies comparing results from separate and joint modelling analyses of survival and longitudinal outcomes were undertaken by \cite{guo_2004}. The simplest formulation possible is the {\it shared random effects model}, where the shared sub-predictor consists of random effects. Choosing a random intercept and slope model, the resulting joint model looks like this:
\begin{eqnarray}
y_{ij} &=& \eta_{\text{l}}(x_{ij}) + \gamma_{0i} + \gamma_{1i}t_{ij}  + \varepsilon_{ij},\nonumber\\
\lambda(t|\alpha,\gamma_{0i}, \gamma_{1i}, x_i) &=& \lambda_0(t)\exp(\eta_{\text{s}}(x_i) + \alpha(\gamma_{0i} + \gamma_{1i}t) ),\nonumber\\
\label{shared_random_effect}
\end{eqnarray}
where the $y_{ij}$ are the longitudinal measurements recorded per individual $i, i = 1,\ldots, N$ at time points $t_{ij}$ with $j = 1,\ldots, n_i$, where $n_i$ is the number of observations recorded for individual $i$. The hazard function $\lambda(t|\alpha, \gamma_{0i}, \gamma_{1i})$ evaluated at time $t$ is based on the baseline hazard $\lambda_0(t)$. The coefficients $\gamma_{0i}$ and $\gamma_{1i}$ are individual-specific random intercept and slope while  $\eta_{\text{l}}(x_{ij})$  and $\eta_{\text{s}}(x_i)$ are the longitudinal and the survival sub-predictor respectively. Both are functions of independent sets of covariates, possibly varying over time in the longitudinal sub-predictor. The association parameter $\alpha$ quantifies the relationship between the two parts of the model. This type of model has been used in many biomedical settings, see for example \cite{gao_2004} or \cite{lei_2007}. For a critical review on shared random effects models in multivariate joint modelling see \cite{Fieuws_2004}. Many extensions have been suggested for model (\ref{shared_random_effect}) as well as a general approach with a universal shared sub-predictor. \cite{Ha_Taesung_Lee} used a generalized model for the longitudinal component, while \cite{Li_Elashoff_Li} suggested an approach for robust modelling of the longitudinal part and included competing risks in the survival part. \cite{Chan} recently included binary outcomes modeled by an autoregressive function, while the model proposed by \cite{armero_2016} accounts for heterogeneity between subjects, serial correlation, and measurement error. Interpretation of the separate parts of the model was simplified by the work of \cite{efendi_2013}. A computationally less burdensome approach has recently been suggested by \cite{barret_2015}. For theoretical results on approximation and exact estimation, see \cite{Sweeting_thompson} and for an overview of the development up to 2004 see \cite{tsiatis}.  Since then \cite{riz_book} has substantially contributed to the research field with the R  add-on package \texttt{JM} \citep[for an introduction to the package see][]{RJM}.

One of the main limitations of classical estimation procedures for joint models in modern biomedical settings is that they are unfeasible for high-dimensional data (with more explanatory variables than patients or even observations). But even in low-dimensional settings the lack of a clear variable selection strategy provides further challenges. In order to deal with these problems, we propose a new inferential scheme for joint models based on gradient boosting \citep{BuhlmannHothorn06}. Boosting algorithms emerged from the field of machine learning and were originally designed to enhance the performance of weak classifiers (\textit{base-learners}) with the aim to yield a perfect discrimination of binary outcomes \citep{nr:freund.schapire:1996}. This was done by iteratively applying them to re-weighted data, giving higher weights to observations that were mis-classified previously. This powerful concept was later extended for use in statistics in order to estimate additive statistical models using simple regression functions as base-learners \citep{friedmanetal2000, friedman_2001}. The main advantages of \textit{statistical boosting} algorithms \citep{boosting_part1} are (i) their ability to carry out automated variable selection, (ii) the ability to deal with high-dimensional $p>n$ data and (iii) that they result in statistical models with the same straightforward interpretation as common additive models estimated via classical approaches \citep{TutzBinder, BuhlmannHothorn06}.

The aim of this paper is the extension of statistical boosting to simultaneously estimate and select multiple additive predictors for joint models in potentially high-dimensional data situations. Our algorithm is based on gradient boosting and cycles through the different sub-predictors, iteratively carrying out boosting updates on the longitudinal and the shared sub-predictor. The model variance, the association parameter and the baseline hazard are optimized simultaneously maximizing the log likelihood. To the best of our knowledge, this is the first statistical-learning algorithm to estimate joint models and the first approach to introduce automated variable selection for potentially high-dimensional data in the framework of joint model.

We apply our new algorithm to data on repeated measurements of lung function in patients with cystic fibrosis in Denmark. Cystic fibrosis is the most common serious genetic disease in white populations. Most patients with cystic fibrosis die prematurely as a result of progressive respiratory disease \citep{dtr2012}. Loss of lung function in cystic fibrosis is accelerated for patients who are infected by pulmonary infections such as {\textit{Pseudomonas aeruginosa}}. However it may also be the case that more rapid lung function decline pre-disposes patients to infection \citep{qvist}. We thus aim to model lung function decline jointly with the onset of infection with {\textit{Pseudomonas aeruginosa}} and select the best covariates from the data set in order to better understand how lung function impacts on risk of infection. This example is suited for a joint modelling approach to provide a better understanding of how lung function influences risk of infection onset, whilst taking into account other covariates such as sex and age that influence both processes.


The remainder of the paper is structured as follows: In the second section we present a short introduction in joint modelling in general and describe how we approach the estimation with a boosting algorithm. In the next section we conduct a simulation study in order to evaluate the ability to
estimate and select effects of various variables for both low-dimensional and high-dimensional data. In the fourth section we apply our approach to cystic fibrosis data with a focus on variable selection. Finally we discuss further extensions of joint models made possible by boosting.

\section{Methods}
%


In this section we describe the model and the associated likelihood as used in the rest of the paper. There are two different dependent variables: a longitudinal outcome and the time of event for the endpoints of interest. The predictor for the longitudinal outcome $y_{it}$ divides into two parts:
$$ y_{ij} = \eta_{\text{l}}(x_{ij}) + \eta_{\text{ls}}(x_i,t_{ij}) + \varepsilon_{ij},$$
where $i = (1, \ldots, N)$ refers to the $i$-th individual, $j = (1,\ldots, n_i)$ to the $j$-th observation and $\varepsilon_{ij}$ is the model error, which is assumed to follow a normal distribution with zero mean and variance $\sigma^2$. The two functions $\eta_{\text{l}}(x_{ij})$ and $\eta_{\text{ls}}(x_i, t_{ij})$, which will be referred to as the longitudinal and the shared sub-predictor, are functions of two separate sets of covariates: $x_{ij}$ are possibly time varying covariates included only in the longitudinal sub-predictor; $x_i$ are covariates varying over individuals yet not over time and are included in the shared sub-predictor. In the setup we are using throughout this paper, the shared sub-predictor will also include a random intercept and slope, denoted by $\gamma_0$ and $\gamma_1$ respectively. 
The shared predictor $\eta_{\text{ls}}(x_i,t_{ij})$ reappears in the survival part of the model:

$$ \lambda(t|\alpha,\eta_{\text{ls}}(x_i, t)) = \lambda_0(t)\exp(\alpha\eta_{\text{ls}}(x_i, t) ), $$

where the baseline hazard $\lambda_0(t)$  is chosen to be constant ($\lambda_0(T_i)=\lambda_0$) in this paper. The sub-predictor with subscript 'ls' refers to both the longitudinal and survival part of the model, whereas it is assumed that the covariates in $\eta_\text{l}(x_{ij})$ only have an impact on the longitudinal structure. The relation between both parts of the model is quantified by the association parameter $\alpha$.
Consequently, the two predictor equations can be summarized in the joint likelihood:
\begin{equation}
\prod_{i=1}^{N}\left\{ \prod_{j=1}^{n_i}f(y_{ij}|\eta_{\text{l}}(x_{ij}), \eta_{\text{ls}}(x_{i},t_{ij}),\sigma^2)
   \right\}f(T_i,\delta_i| \eta_{\text{ls}}(x_i, T_i),\alpha, \lambda_0),
	\label{like}
	\end{equation}
where $T_i$ is the observed time of event for individual $i$ and where the distribution for the longitudinal part is the Gaussian error distribution.
The parameter $\delta_i$ is the censoring indicator for the $i$-th individual, taking the value $0$ in the case of censoring and $1$ in the case of an event. The likelihood for the survival part is:
\begin{equation*}
f(T_i, \delta_i|\alpha, \eta_{\text{ls}i}(T_i), \lambda_0)=\left[\lambda_0(T_i)\exp(\alpha(\eta_{\text{ls}i}(T_i)))\right]^{\delta_i}\exp\left[-\lambda_0\int_0^{T_i}\!\exp(\alpha\eta_{\text{ls}}(x_i,u))\ \text{d}u\right] \ .
\end{equation*}

\subsection{Component-wise gradient boosting}\label{sec:methodsboost}

\begin{algorithm}[t]
    \caption{Component-wise gradient boosting}\label{alg:gradboost}
		\begin{algorithmic}
		\STATE Initialize the additive predictor with a starting value, e.g. $\hat{\eta}^{[0]} := (0)_{i = 1,\ldots,n}$. Specify a set of base-learners $h_1(x_1), \ldots, h_p(x_p)$.
     \FOR{$m = 1$ to $m_{\text{stop}}$}

     \STATE \textbf{Fit the negative gradient} \\
								\bindent
								\STATE Increase iteration counter $m := m + 1$
                                \STATE Compute the negative gradient vector $\boldsymbol{u}^{[m]}$ of the loss function evaluated at current $\eta$:
                                    $$ \boldsymbol{u}^{[m]} = \left(u_i^{[m]}\right)_{i = 1,\ldots,n} = \left( - \left. \frac{\partial}{\partial \eta} \rho(y_i, \eta_i) \right|_{\eta_i = \hat{\eta}^{[m-1]}(x_i)} \right)_{i = 1,\ldots,n}$$

                                \STATE Fit the negative gradient vector $\boldsymbol{u}^{[m]}$ separately to every base-learner:
								 $$\boldsymbol{u}^{[m]} \xrightarrow{\text{base-learner}} \hat{h}_{j}^{[m]}(x_j)\quad \text{for } j=1,\ldots,p.$$						
								\eindent
					
	\STATE \textbf{Update one component}\\
								\bindent
                                \STATE Select the component $j^*$ that best fits $\boldsymbol{u}^{[m]}$:
                                  $$ j^* = \underset{1 \leq j \leq p}{\operatorname{argmin}}\sum_{i=1}^n (u_{i}^{[m]} - \hat{h}_{j}^{[m]}(x_j))^2 \  $$
                                 \STATE Update the additive predictor with this base-learner fit,
                                 $$\hat{\eta}^{[m]}= \hat{\eta}^{[m-1]} + \nu \cdot \hat{h}^{[m]}_{j^*}(x_{j^*})$$
                                 where $\nu$ is the step-length, a typical value in practice is 0.1 \citep{mboost_tut}.
								\eindent
											
		\ENDFOR{  $m =  m_{\text{stop}}$}
		\end{algorithmic}
\end{algorithm}

In the following section we briefly highlight the general concept of boosting before we will adapt it to the class of joint models.  The basic idea of statistical boosting algorithms is to find an additive predictor $\eta$ for a statistical model that optimizes the expected loss regarding a specific loss function $\rho(y_i, \eta_i)$. The loss function describes the type of regression setting and denotes the discrepancy between realizations $y_i$ and the model $\eta_i = \eta(x_i)$. The most typical example for a loss function is the $L_2$ loss for classical regression of the expected value.

For a given set of observation $y_1, \ldots, y_n$, the algorithm searches for the best solution to minimize the empirical loss (often referred to as \textit{empirical risk}) for this sample:
\begin{eqnarray*}
	\hat{\eta} = \underset{\eta}{\operatorname{argmin}} \ \frac{1}{n} \sum_{i = 1}^n \left( \rho(y_i, \eta(x_i)) \right) \ .
\end{eqnarray*}
In case of the classical $L_2$ loss, the empirical risk simply refers to the mean squared error. While there exist different approaches for statistical boosting \citep{boosting_part1}, we will focus in this work on component-wise gradient boosting \citep{BuhlmannHothorn06}. The main concept is to iterative apply base-learners $h_1(x_1), \ldots, h_p(x_p)$, which are typically simple regression type functions that use only one component of the predictor space (i.e., one covariate $x_j$). The base-learners are fitted one-by-one (component-wise), not on the observations $y_1,\ldots,y_n$ but on the negative gradient vector $u_1,\ldots,u_n$ of the loss
\begin{eqnarray*}
 \boldsymbol{u}^{[m]} = \left(u_i^{[m]}\right)_{i = 1,\ldots,n} = \left( - \left. \frac{\partial}{\partial \eta} \rho(y_i, \eta_i) \right|_{\eta_i = \hat{\eta}^{[m-1]}(x_i)} \right)_{i = 1,\ldots,n} \ ,
\end{eqnarray*}
at the $m$-th iteration step. In case of the $L_2$ loss, this means that the base-learners in iteration $m$ are actually fitted on the residuals from iteration $m-1$. 

In classical boosting algorithms from the field of machine learning, base-learners are simple classifiers for binary outcomes. In case of statistical boosting, where the aim is to estimate additive predictors, the base-learners itself are regression models: The base-learner $h_j(x_j)$ represents the potential partial effect of variable $x_j$ on the outcome. Examples are simple linear models ($h_j(x_j) = \beta_j \cdot x_j$) or smooth non-parametric terms estimated via splines.

In every boosting iteration, only the best-performing base-learner is selected to be updated in the additive predictor (see Algorithm~\ref{alg:gradboost}). Typically, one base-learner is used for each variable. The specification of the base-learner defines the type of effect the variable is assumed to have in the predictor. For linear effects one could for example use simple linear models \citep{Buehlmann_2006} or P-splines for non-linear effects \citep{Schmid:Hothorn:boosting-p-Splines}.

The stopping iteration $m_{\text{stop}}$ is the main tuning parameter as it controls variable selection and shrinkage of effect estimates. If the algorithm is stopped before convergence, base-learners that have never been selected for the update are effectively excluded from the final model. Higher numbers of $m_{\text{stop}}$ hence lead to larger, more complex models while smaller numbers lead to sparser models with less complexity. In practice, $m_{\text{stop}}$ is often selected via cross-validation or resampling methods, by selecting the value that leads to the smallest empirical risk on test data \citep{mboost_tut}.

For theoretical insights on the general concept of boosting algorithms, we refer to the work of \cite{zhang2005boosting} who studied the numerical convergence and consistency with different loss functions. For the $L_2$-loss, \cite{Buehlmann_2006} proved the consistency of gradient boosting with simple linear models as base-learners in the context of high-dimensional data \citep[cf.,][]{hepp2016approaches}.

\subsection{Boosting for multiple dimensions}

\begin{algorithm}[t]
	\caption{Component-wise gradient boosting for multiple dimensions}\label{alg:gamboostLSS}
	\begin{algorithmic}
		\STATE Initialize additive predictors for parameters $\theta_1,\ldots, \theta_K$ with starting values, e.g. $\hat{\eta}_{\theta_k}^{[0]} := (0)_{i = 1,\ldots,n}$ for $k = 1,\ldots,K$. For each of these $K$ dimensions, specify a set of base-learners $h_{k1}(\cdot), \ldots, h_{kp_k}(\cdot)$, where $p_k$ is the cardinality of the set of base-learners specified for $\theta_k$, typically this refers to the number of candidate variables. Initialize nuisance parameter $\hat{\phi}^{[0]}$ with offset.
		
		\FOR{$m = 1$ to $m \geq m_{\text{stop},k}$ for all $k = 1,\ldots,K$}
		\FOR{$k = 1$ to $K$}
		\STATE  \textbf{if} $ m > m_{\text{stop},k}$ set $\hat{\eta}_{\theta_k}^{[m]} := \hat{\eta}_{\theta_k}^{[m-1]}$ and skip this iteration.

		\STATE \textbf{Fit the negative gradient} \\
		\bindent
		\STATE Compute the negative gradient of the loss evaluated at current $\boldsymbol{\eta}^{[m-1]}_i  = (\hat{\eta}^{[m-1]}_{\theta_1},\ldots,\hat{\eta}^{[m-1]}_{\theta_K})$ and $\phi = \hat{\phi}^{[m-1]}$:
		$$ \boldsymbol{u}_k^{[m]} =  \left( -  \frac{\partial}{\partial \eta_{\theta_k}} \rho(y_i, \boldsymbol{\eta}^{[m-1]}_i, \phi) \right)_{i = 1,\ldots,n}$$
		
		\STATE Fit the negative gradient vector $\boldsymbol{u}_k^{[m]}$ separately to every base-learner defined for dimension $k$:
		$$\boldsymbol{u}_k^{[m]} \xrightarrow{\text{base-learner}} \hat{h}_{kj}^{[m]}(\cdot)\quad \text{for } j=1,\ldots,p_k.$$						
		\eindent
		
		\STATE \textbf{Update one component}\\
		\bindent
		\STATE Select the component $j^*$ that best fits $\boldsymbol{u}_k^{[m]}$:
		$$ j^* = \underset{1 \leq j \leq p_k}{\operatorname{argmin}}\sum_{i=1}^n (u_{ik}^{[m]} - \hat{h}_{kj}^{[m]}(\cdot))^2 \  $$
		\STATE Update the additive predictor with this base-learner fit:
		$$\hat{\eta}_{\theta_k}^{[m-1]} := \hat{\eta}_{\theta_k}^{[m-1]} + \nu \cdot \hat{h}^{[m]}_{kj^*}(\cdot)$$
		\STATE Set $\hat{\eta}_{\theta_k}^{[m]} = \hat{\eta}_{\theta_k}^{[m-1]} $
		\eindent
		
		\ENDFOR{  $k =  K$}
		
		\STATE \textbf{Update nuisance parameter} \\
		\bindent
		\STATE \textbf{if} $\phi$ is a nuisance parameter, that should not be modelled, find the optimal scalar:
		
		$$\hat{\phi}^{[m]} = \underset{\phi}{\operatorname{argmin}} \sum_{i=1}^n \rho(y_i, \boldsymbol{\hat{\eta}}^{[m]}_i , \phi)  $$
		
		\eindent

		\ENDFOR{  $m \geq  m_{\text{stop},k}$ for all $k= 1,\ldots,K$}
	\end{algorithmic}
\end{algorithm}

The general concept of component-wise gradient boosting was later extended to numerous regression settings \citep[for a recent overview, see][]{boosting_part2}. Some of these extensions focused on loss functions that can be optimized with respect to multiple dimensions simultaneously \citep{Schmid:Multidim:2010, gamboostlss:2012}. This can refer either to regression settings where multiple distribution parameters $\theta_1, \ldots,  \theta_K$ should be modelled, e.g., $\boldsymbol{\eta_{\theta}} = (\eta_{\theta_1}, \dots, \eta_{\theta_K})$ like in distributional regression \citep{gamlss}, or settings where in addition to the main model $\eta$ some nuisance parameter (e.g., a scale parameter $\phi$ for negative binomial regression) should be optimized simultaneously. 

Boosting the latter model can be achieved by first carrying out the classical gradient-fitting and updating steps for the additive predictor (see Algorithm~\ref{alg:gradboost}) and second by updating the nuisance parameter $\phi$, both in each iteration step. Updating the nuisance parameter is done by optimizing the loss function with respect to $\phi$, keeping the current additive predictor fixed:
\begin{eqnarray}
  \hat{\phi}^{[m]} = \underset{\phi}{\operatorname{argmin}} \sum_{i=1}^n \rho(y_i, \hat{\eta}^{[m]}_i , \phi)  \ .
\end{eqnarray}

A typical example for a regression setting where various parameters $\boldsymbol{\theta} = (\theta_1,\ldots,\theta_K)$ should be modeled simultaneously by $K$ additive predictors are generalized additive models for location, scale and shape \citep[GAMLSS, ][]{gamlss}. The idea is to model all parameters of a conditional distribution by their own additive predictor and own associated link function. This involves not only the location (e.g., the mean), but also scale and shape parameters (e.g., variance, skewness). 

In case of boosting GAMLSS, the algorithms needs to estimate $K$ statistical models simultaneously. This is achieved by incorporating an outer loop that circles through the different distribution parameters, always carrying out one boosting iteration and updating them one by one (see Algorithm~\ref{alg:gamboostLSS} for details). As a result, the algorithm can provide intrinsic shrinkage and variable selection for $K$ models simultaneously.

\subsection{Boosting Joint Models}

The algorithm we suggest for estimating the sub-predictors for joint modelling is based on the boosting algorithm for multiple dimensions as outlined in the previous part of this section, but it differs in a range of aspects from Algorithm ~\ref{alg:gamboostLSS}. The main reason for those differences is that the additive predictors for the two dependent variables (the longitudinal outcome and the time-to-event) are neither entirely different nor completely identical. While the longitudinal outcome is modelled by the set of the two sub-predictors $\eta_{\text{l}}$ and $\eta_{\text{ls}}$, the survival outcome in our model is solely based on the shared sub-predictor $\eta_{\text{ls}}$ and the corresponding association parameter $\alpha$. It is hence not possible to construct the algorithm circling between the two outcomes $y_{ij}$ and $\lambda(t)$ as for the components $\theta_k$ in the preceding section. Translating the joint likelihood to one empirical loss function poses difficulties, since the losses of the different observations differ highly in their dimensionality. We therefore suggest an updating scheme at predictor stage (i.e. $\eta_{\text{l}}(x_{ij})$ and $\eta_{\text{ls}}(x_i, t_{ij})$) rather than at the level of the dependent variables.  More specifically, we define an outer loop that cycles in every boosting iteration through the following three steps:

\begin{enumerate}
\item[{}]\textbf{(step1)} update $\eta_{\text{l}}(x_{ij})$ in a boosting iteration
\item[{}]\textbf{(step2)} update $\eta_{\text{ls}}(x_{i}, t_{ij})$ in a boosting iteration
\item[{}]\textbf{(step3)} update $\alpha$, $\lambda_0$ and $\sigma^2$ by maximizing the likelihood.
\end{enumerate}

 We omit the arguments of the sub-predictors to ensure readability in the following sections. The longitudinal sub-predictor will hence be denoted by the indexed values: $\eta_{\text{l}ij}$ for the longitudinal and $\eta_{\text{ls}ij}$ for the shared sub-predictor. The derivation of the parts necessary for the three steps will be described in the following, with exact calculations in Appendix A. In the tradition of the \textit{shared random effects models} we will consider a setup including random effects in the shared sub-predictor and furthermore allowing for a fixed effect linear in time. The structure of $\eta_{\text{ls}}$ is hence
\begin{eqnarray*}
\eta_{\text{ls}ij} = \tilde{\eta}_{\text{ls}i} + \beta_t t_{ij}+\gamma_{0i}+\gamma_{1i}t_{ij},
\end{eqnarray*}
where $\tilde{\eta}_{\text{ls}i}$ can include various different non time dependent covariate functions.

\begin{algorithm}[ht!]
    \caption{Component-wise gradient boosting for Joint Models}\label{alg:boostJM}
		\begin{algorithmic}
		\STATE \textbf{Initialize} $\eta_{\text{l}}$ and $\eta_{\text{ls}} $with starting values, e.g. $\hat{\eta}_{\text{l}}^{[0]} = \hat{\eta}_{\text{ls}}^{[0]}:= (0)_{i = 1,\ldots,n}$. Specify a set of base-learners $(h_{\text{l}1}(\cdot), \ldots, h_{\text{l}p_{\text{l}}}(\cdot))$ and $(h_{\text{ls}1}(\cdot), \ldots, h_{\text{ls}p_\text{ls}}(\cdot))$ where $p_{\text{l}}$ and $p_{\text{ls}}$ are the cardinality of the set of base-learners. Initialize baseline hazard  and association parameter  with offset values e.g., $\lambda_0^{[0]} := 0.1$  and $\alpha^{[0]} := 0.1$.

      \FOR{$m = 1$ to $m \geq m_{\text{stop,l}}$ and $m \geq m_{\text{stop,ls}}$}
           \STATE \textbf{step1: Update exclusively longitudinal predictor in a boosting step} \\
								\bindent
                                \STATE  \textbf{if} $ m > m_{\text{stop,l}}$ set $\hat{\eta}_{\text{l}}^{[m]} := \hat{\eta}_{\text{l}}^{[m-1]}$ and skip this step.

								\STATE
                                        Compute $\boldsymbol{u}^{[m]}_{\text{l}}$ as
																				 \begin{small}
                                        $$\boldsymbol{u}^{[m]}_{\text{l}}= \left(u^{[m]}_{\eta_{\text{l}i}}\right)_{i=1,\ldots,n} = \frac{1}{\sigma^2}\left(y_i - \hat{\eta}^{[m-1]}_{\text{ls}i} - \hat{\eta}^{[m-1]}_{\text{l}i} \right)_{i = 1,\ldots,n}.$$
																				 \end{small}
								\STATE Fit the negative gradient vector   $\boldsymbol{u}^{[m]}_{\text{l}}$ separately to every base-learner specified for $\eta_{\text{l}}$:
								                              \begin{small}
								      	 $$\boldsymbol{u}_{\text{l}}^{[m]} \xrightarrow{\text{base-learner}} \hat{h}_{\text{l}j}^{[m]}(\cdot)\quad \text{for } j=1,\ldots,p_{\text{l}}.$$	
												                              \end{small}
								\STATE Select the component $j^*$ that best fits $\boldsymbol{u}_{\text{l}}^{[m]}$:
                                  $$ j^* = \underset{1 \leq j \leq p_{\text{l}}}{\operatorname{argmin}}\sum_{i=1}^n (u_{\text{l}i}^{[m]} - \hat{h}_{\text{l}j}^{[m]}(\cdot))^2 \ $$

								\STATE and update this component: $\hat{\eta}_{\text{l}}^{[m]}= \hat{\eta}_{\text{l}}^{[m-1]} + \nu \cdot \hat{h}^{[m]}_{\text{l}j^*}(\cdot)$ .
								\eindent
						
                            \STATE \textbf{step2: Update joint predictor in a boosting step}\\
								\bindent
							    \STATE  \textbf{if} $ m > m_{\text{stop,ls}}$ set $\hat{\eta}_{\text{ls}}^{[m]} := \hat{\eta}_{\text{ls}}^{[m-1]}$ and skip this step.
                            	\STATE Compute $\boldsymbol{u}^{[m]}_{\text{ls}}$ as
                                \begin{small}
																\begin{eqnarray*}
                                 &&\boldsymbol{u}^{[m]}_{\text{ls}}  = \\
                                 &&\left( \frac{y_{ij} - \eta^{[m]}_{\text{l}i} -\eta^{[m-1]}_{\text{ls}i}}{\sigma^2} +\delta_i\alpha^{[m-1]}-\frac{\lambda_0^{[m-1]}\left(\exp\left(\alpha^{[m-1]}\eta^{[m-1]}_{\text{ls}i}\right)-\exp\left(\alpha^{[m-1]}\eta^{[m-1]}_{\text{ls}i-t}\right)\right)}{\beta^{[m-1]}_t + \gamma^{[m-1]}_{1i}} \right)_{i = 1\ldots,N, j= 1,\ldots,n_i}
                                 \end{eqnarray*}
																\end{small}
								\STATE Fit the negative gradient vector   $\boldsymbol{u}^{[m]}_{\text{ls}}$ separately to every base-learner specified for $\eta_{\text{ls}}$:
								                              \begin{small}
								      	 $$\boldsymbol{u}_{\text{ls}}^{[m]} \xrightarrow{\text{base-learner}} \hat{h}_{\text{ls}j}^{[m]}(\cdot)\quad \text{for } j=1,\ldots,p_{\text{ls}}.$$								
												                              \end{small}
								\STATE Select the component $j^*$ that best fits $\boldsymbol{u}_{\text{ls}}^{[m]}$
								                              \begin{small}
                                  $$ j^* = \underset{1 \leq j \leq p_{\text{ls}}}{\operatorname{argmin}}\sum_{i=1}^n (u_{\text{ls}i}^{[m]} - \hat{h}_{\text{ls}j}^{[m]}(\cdot))^2 \  ,$$
                              \end{small}
                                \STATE and update this component $\hat{\eta}_{\text{ls}}^{[m]}= \hat{\eta}_{\text{ls}}^{[m-1]} + \nu \cdot \hat{h}^{[m]}_{\text{ls}j^*}(\cdot)$.
							\eindent
						\STATE \textbf{step3: Update $\sigma^2$, $\alpha$ and $\lambda_0$ by maximizing the likelihood}\\
								\bindent
								\STATE
								                              \begin{small}
								${\sigma^2}^{[m]} := \underset{\sigma^2}{\operatorname{argmax}} \    \prod_{i,j} f\left(y_{ij}, T_i, \delta_i|\alpha^{[m-1]},\eta^{[m]}_{\text{ls}i}(\cdot), \eta^{[m]}_{\text{l}i}(\cdot),\lambda_0^{[m-1]}, \sigma^2\right)$
								                              \end{small}
							    \STATE  \textbf{if} $ m > m_{\text{stop,ls}}$ set $\hat{\alpha}^{[m]} := \hat{\alpha}^{[m-1]}$ $\hat{\lambda}^{[m]} := \hat{\lambda}^{[m-1]}$ and skip this step:
               	\STATE                                \begin{small}
								$ (\alpha^{[m]}, \lambda_0^{[m]}) := \underset{\alpha, \lambda_0}{\operatorname{argmax}} \    \prod_{i,j} f\left(y_{ij}, T_i, \delta_i|\alpha,\eta^{[m]}_{\text{ls}i}(\cdot), \eta^{[m]}_{\text{l}i}(\cdot),\lambda_0, {\sigma^2}^{[m]}\right)$
								                              \end{small}
								\eindent
						
		\ENDFOR{ $m \geq m_{\text{stop,l}}$ and $m \geq m_{\text{stop,ls}}$}
		\end{algorithmic}
\end{algorithm}

\paragraph{(step1) Boosting the exclusively longitudinal predictor:} 
The optimization problem of the longitudinal predictor $\eta_{\text{l}}(\cdot)$ is straight-forward, since it is basically the same as for a quadratic loss function. The gradient vector at iteration $m$ hence consists of the differences (residuals) from the iteration before:
\begin{eqnarray*}
u_{\text{l}i}^{[m]} =  \frac{1}{\sigma^2}\left(y_{ij} - \eta_{\text{ls}ij}^{[m-1]} -\eta_{\text{l}ij}^{[m-1]}\right),
\end{eqnarray*}

The vector $\boldsymbol{u}_{\text{l}}$ has the same length as the vector $\boldsymbol{y}$, including all longitudinal measurements before the event/censoring. Note that there is a slight difference to the gradient vector for classical Gaussian regression: the variance $\sigma^2$ has to be included to ensure that the weighting between the two parts of the model is correct.

\paragraph{(step2) Boosting the joint predictor:} 
The optimization problem for the joint predictor $\eta_{\text{ls}}(\cdot)$ is more complex. To account for the different scales of the different outcomes, we construct a loss function in such a way that there is an entry for each individual rather than for each observation. This leads to $n$ entries in the vector of loss functions that consist of two parts for each individual $i$. The entries of the loss vector hence are:
\begin{eqnarray*}
\frac{1}{2\sigma^2} \left(y_{ij} - \eta_{\text{ls}ij}- \eta_{\text{l}ij}\right)^2+ (-\log(f(T_i, \delta_i|\alpha, \eta_{\text{ls}ij} \lambda_0))),
\label{eq:loss}
\end{eqnarray*}

with $i=1\ldots n, j= 1,\ldots,n_i$. The resulting entries of the gradient vector  $\boldsymbol{u}_{\text{ls}}$ at iteration $m$, after the update of $\eta_{\text{l}}(\cdot)$, are:
\begin{eqnarray*}
u_{\text{ls}ij}^{[m]} =  \frac{1}{\sigma^2}\left(y_{ij} - \eta_{\text{l}ij}^{[m]} -\eta_{\text{ls}ij}^{[m-1]}\right) +\delta_i\alpha^{[m-1]}-\frac{\lambda_0^{[m-1]}(\exp(\alpha^{[m-1]}\eta^{[m-1]}_{\text{ls}ij})-\exp(\alpha^{[m-1]}\eta^{[m-1]}_{\text{ls}ij-t}))}{\beta^{[m-1]}_t + \gamma^{[m-1]}_{1i}},
\end{eqnarray*}
where $\beta_t$ and $\gamma_1$ are the coefficients for fixed time-effect and random slope in the shared sub-predictor $\boldsymbol{\eta}_{\text{ls}}$ and $\boldsymbol{\eta}_{\text{ls}-t}$ is the part of the sub-predictor not depending on time. For the complete derivation of this result see Appendix A.

\paragraph{(step3) Estimating the association parameter, the baseline hazard and the variance:}

In order to estimate $\alpha$, $\lambda_0$ and $\sigma^2$, we maximize the likelihood (\ref{like}) simultaneously in every iteration.
\begin{eqnarray*}
	 (\alpha^{[m]}, \lambda_0^{[m]}, {\sigma^2}^{[m]}) := \underset{\alpha, \lambda_0, \sigma^2}{\operatorname{argmax}} \     f\left(y_{ij}, T_i, \delta_i|\alpha,\eta^{[m]}_{\text{ls}ij}, \eta^{[m]}_{\text{l}ij},\lambda_0, \sigma^2 \right)
\end{eqnarray*}

This third step is carried out after every boosting iteration for the joint predictor (step 2), even if boosting the longitudinal predictor was already stopped and the corresponding steps were hence skipped.\\

The complete proposed algorithm for boosting JM is presented as Algorithm~\ref{alg:boostJM} and its implementation is provided with the new R add-on package \texttt{JMboost} which source code is hosted openly on\\ \texttt{http://www.github.com/mayrandy/JMboost}. \\

\paragraph{Model tuning:}
Tuning of the algorithm is similar to the classical boosting algorithms for multidimensional loss functions: In general, both the step-length $\nu$ and the different stopping iterations $m_{\text{stop}}$ have an influence on the variable selection properties, convergence speed and the final complexity of the additive predictors. However, in practice, there exists a quasi-linear relation between the step-length and the needed number of boosting iterations \citep{Schmid:Hothorn:boosting-p-Splines}. As a result, it is often recommended to use a fixed small value of $\nu = 0.1$ for the step-length and optimize the stopping iteration instead \citep{mayr2012importance,mboost_tut}.

In case of boosting JM, where two additive predictors are fitted and potentially two boosting updates are carried out in each iteration of the algorithms, it is hard to justify why both predictors should be optimal after the same number of boosting iterations (i.e. $m_{\text{stop,l}} =  m_{\text{stop,ls}}$). This special case was referred to as \textit{one-dimensional early stopping} in contrast to the more flexible \textit{multi-dimensional early stopping}, where instead of one single $m_{\text{stop}}$ value a vector of stopping iterations $(m_{\text{stop,l}} , m_{\text{stop,ls}})$ has to be optimized via a grid \citep{gamboostlss:2012}.  This computationally burdensome issue is necessary to allow for different levels of complexity for the additive predictors. The range of the grid (minimum and maximum $m_{\text{stop}}$) has to be specified \textit{ad-hoc}, however it might be necessary to adapt it based on the results (\textit{adaptive} grid search).

For a more detailed discussion on how to select this grid in practice, we refer to \cite{gamboostLSS_tut}. Finally, the combination of stopping iterations from the grid is selected, which yields the smallest empirical risk on test-data (e.g., via cross-validation or resampling procedures).


\section{Simulations}
To evaluate the performance of the new boosting algorithm, a generic simulation setup was created to mimic a typical joint modelling situation, and will be described in the following. The technical details for the complete simulation algorithm are explained in depth in Appendix B, the underlying R-Code to reproduce the results is included in the Supporting Information.

\subsection{Simulation Setup}

The purpose of the simulation study is threefold:
\begin{itemize}
\item[(i)] evaluate estimation accuracy,
\item[(ii)] test variable selection properties, and
\item[(iii)] determine the appropriate stopping iterations $m_{\text{stop,l}}$ and $m_{\text{stop,ls}}$.
\end{itemize}

The evaluation of estimation accuracy has to be done for both, the coefficients of the sub-predictors as well as the association parameter, which plays a crucial role in the interpretation of the model.
Three different setups were chosen to give insight in all three questions from different angles. We will first describe the basis for all three models and then point out the differences in the simulation concepts. All three models are based on the following sub-predictors:
\begin{eqnarray*}
\eta_{\text{l}ij} = \boldsymbol{x}^{\top}_{\text{l}ij}\boldsymbol{\beta}_{\text{l}} \quad \text{and}\quad \eta_{{\text{ls}}ij} = \boldsymbol{x}^{\top}_{{\text{ls}}i}\boldsymbol{\beta}_{\text{ls}} + \beta_t t_{ij} + \gamma_{0i} + \gamma_{1i}t_{ij}.
\end{eqnarray*}

The matrices $\boldsymbol{X}_{\text{l}}$ and $\boldsymbol{X}_{\text{ls}}$ are the collections of the standardised covariates, $\beta_{\text{l}}=(2,1,-2, \boldsymbol{0})^{\top}$ and $\beta_{\text{ls}}=(1,-2, \boldsymbol{0})^{\top}$ the corresponding linear effects with sub vectors $\boldsymbol{0}$ of different lengths, $\beta_t = 1$ is the impact of time $\boldsymbol{t}$, $\gamma_0$ the random intercept and $\gamma_1$ the random slope. In all three setups $N=500$ individuals were generated. For each individual, we drew five time points which were reduced to less than five in cases with events before the last observation. If there was no incident for any of the simulated individuals the number of observations would thus be $2500$. However the simulations are constructed in a way that this case never occurs. The first of the three setups (S1) was constructed to mimic the data situation of the application in Section~\ref{sec:data} more closely and to thus better demonstrate the ability of the algorithm to perform variable selection in a lower dimension. S1 had four non informative covariates in each predictor. In the second simulation setup (S2) we used $600$ non informative covariates, $300$ for each sub-predictor. In this case the number of covariates exceeds the number of individuals. In the third of the three setups (S3) we chose the number of non informative covariates to be $2500$ over both sub-predictors, i.e. $1250$ each, the theoretical maximum number of observations is hence exceeded by the number of covariates. Please note that we first simulated a genuine informative part of the model, and afterwards drew the non informative covariates for each setup individually. The three setups are hence based on the same data and the informative covariates of the three models are the same. In all three setups the association parameter was chosen to be $0.5$.


\subsection{Results}

We ran $100$ models of each setup; results are summarized in Table~\ref{beta:tab} and will be described in detail in the following.
\begin{center}
\begin{table}
\begin{small}
\begin{tabular}{|r|rrrr|cccc|rr|}
\hline
Setup&$\hat{\beta}_{\text{l}}\left(\text{sd}\right)$&$\beta_{\text{l}}$&$\hat{\beta}_{\text{ls}}\left(\text{sd}\right)$&$\beta_{\text{ls}}$&TP$_{\text{l}}$&FP$_{\text{l}}$&TP$_{\text{ls}}$&FP$_{\text{ls}}$ &$\hat{\alpha}\left(\text{sd}\right)$&$\alpha$\\
\hline
&$2.025 (0.015)$&$2$&$0.989 (0.008)$&$1$&&&$1.00$&&&\\
S1&$0.998 (0.009)$&$1$&$-1.991 (0.009)$&$-2$&$1.00$&$0.782$&$1.00$&$0.015$&$0.520(0.076)$&$0.5$\\
&$-1.998 (0.008)$&$-2$&$0.929 (0.031)$&$1$&$1.00$&&$1.00$&&&\\
&&&&&&&&&&\\
&$2.031 (0.018)$&$2$&$0.987 (0.009)$&$1$&&&$1.00$&&&\\
S2&$0.984 (0.011)$&$1$&$-1.99(0.009)$&$-2$&$1.00$&$0.004$&$1.00$&$0.008$&$0.521(0.077) $&$0.5$\\
&$-1.986 (0.011)$&$-2$&$0.915 (0.038)$&$1$&$1.00$&&$1.00$&&&\\
&&&&&&&&&&\\
&$2.037(0.018)$&$2$&$0.985(0.009)$&$1$&&&$1.00$&&&\\
S3&$0.982(0.012)$&$1$&$-1.986(0.009)$&$-2$&$1.00$&$0.001$&$1.00$&$0.002$& $0.522 (0.077)$&$0.5$\\
&$-1.985(0.011)$&$-2$&$0.898(0.037)$&$1$&$1.00$&&$1.00$&&&\\
\hline
\end{tabular}
\end{small}
\caption{\label{beta:tab} Estimates for the coefficients and selection proportions of the variables in the three simulation setups for $100$ simulation runs. TP stands for true positive and indicates the selection rate for each informative variable individually. Intercepts are in the model automatically, hence no selection frequency is reported. FP stands for false positive and denotes the overall selection rate for non-informative variables per model.}
\end{table}
\end{center}

\begin{figure}[t]\centering
\includegraphics[width=\textwidth]{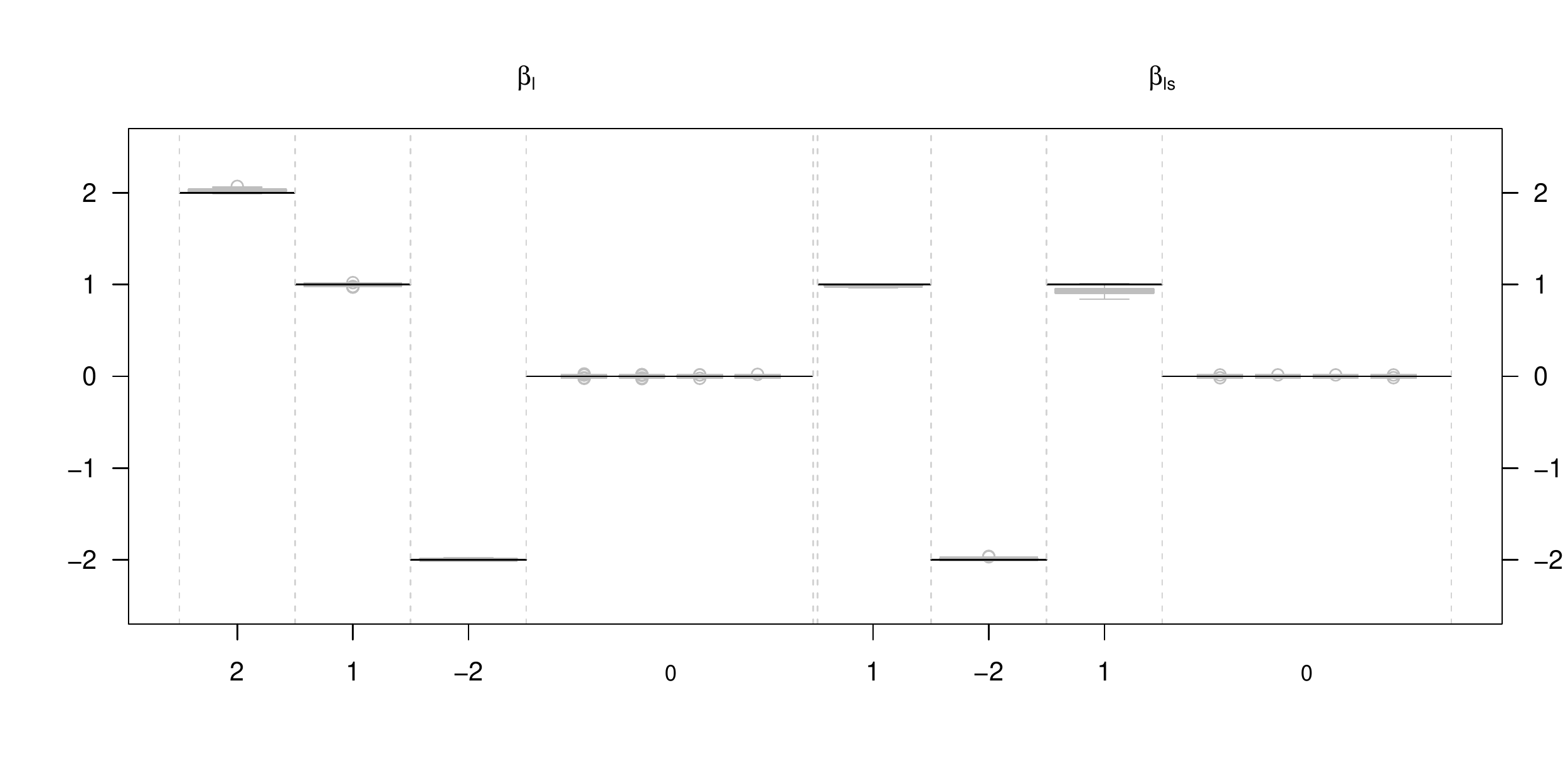}

\caption{\label{beta} Boxplot representing the empirical distribution of the resulting coefficients for simulation setting S1 over 100 simulation runs: on the left side coefficients for the longitudinal sub-predictor are displayed, on the right side coefficients for the shared sub-predictor. The black solid lines indicate the true values. The narrow boxes display the estimation for the effects of the non-informative variables}
\end{figure}

\begin{figure}[!ht]\centering
\includegraphics[width=.5\textwidth]{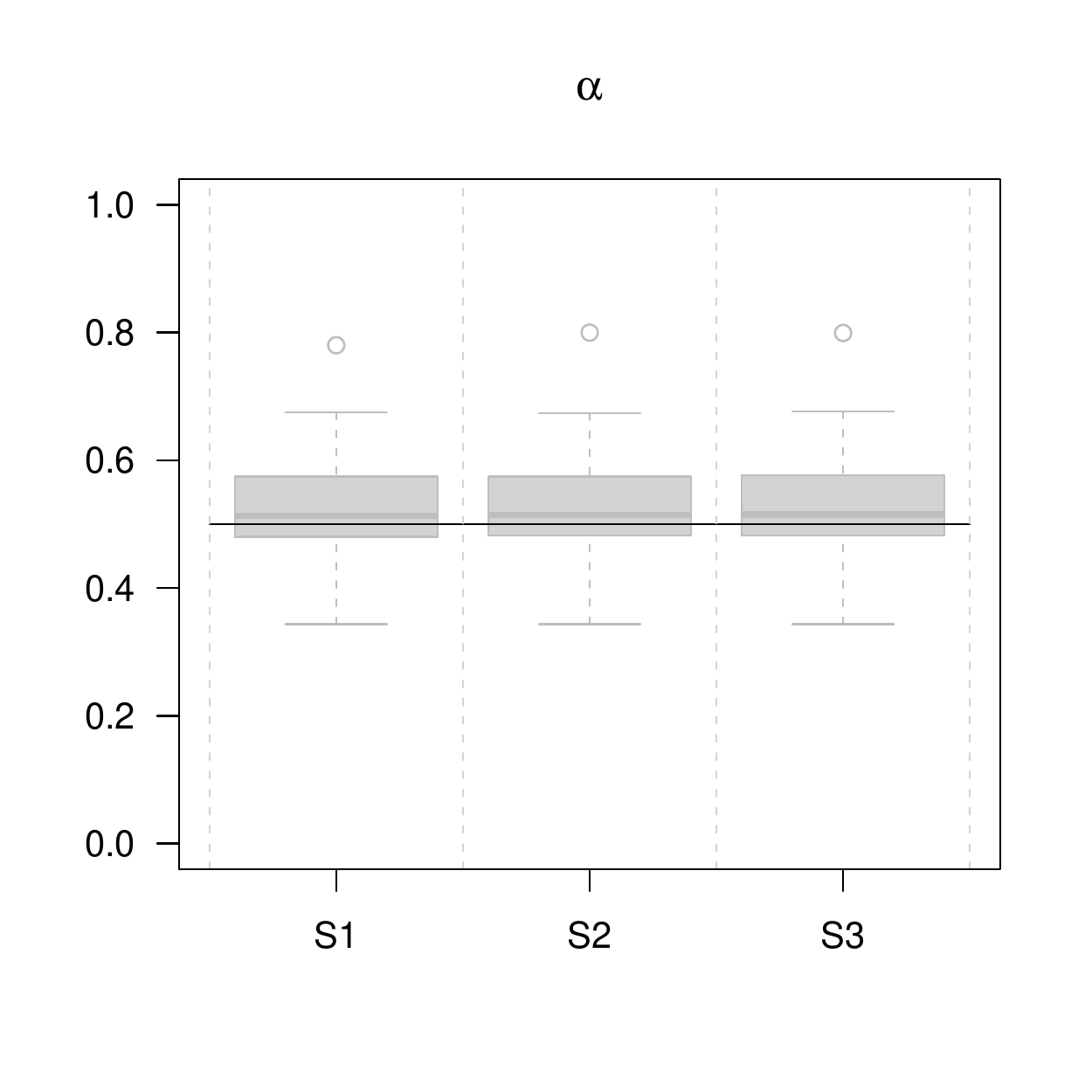}
\caption{\label{alpha} Boxplot of the resulting estimates of the association parameter $\alpha$ over all three simulation setups and 100 simulation runs. The black solid lines indicate the true parameter.}
\end{figure}

\paragraph{(i) Evaluation of estimation accuracy}
As can be seen in Table~\ref{beta:tab} estimation for the coefficients of the sub-predictors were very close to the true values. Standard deviations were small, except for the estimation of the time effect, where the variation between the simulation runs is slightly higher. A graphical illustration is provided in Figure ~\ref{beta}, which displays boxplots of the resulting coefficients for S3 (the results for the other setups only differs slightly). The slight tendency of underestimation can be attributed to the shrinkage behavior of boosting algorithms in general. This underestimation can be observed in all three models in both parts. The estimation of the crucial association parameter $\alpha$ was also very close to the true value in all setups (see the last column of Table~\ref{beta:tab} and Figure~\ref{alpha}). The slight tendency to overestimate $\alpha$ can be contributed to the compensatory behaviour of the association parameter. This is due to the fact that $\alpha$ is estimated via optimisation and hence not subject so shrinkage but adapting directly to the data.

\paragraph{(ii) Variable selection properties}
All informative variables were selected in $100$\% of the runs in all three setups. In S1, more non-informative variables ($76.8\%$) were selected in the longitudinal predictor than in the shared predictor ($1.5\%$). Boosting tends to select a higher percentage of non-informative variables in a small setup, which explains the high selection rate in the longitudinal component. The fact that the shared part is less prone to this greedy behaviour of the algorithm can be contributed to the option to chose the random effect over one of the non-informative variables. In the high-dimensional setups non-informative effects are selected in very few of the runs in both setups. The longitudinal sub-predictor does even better than the shared sub-predictor in this case. There are slight differences, which can be attributed to the smaller number of runs it required (see paragraph below). Overall the selection property works very well for both parts of the model, especially in high-dimensional scenarios.

\paragraph{(iii) Stopping iterations}
The two (possibly different) stopping iterations $m_\text{stop,l}$ and $m_{\text{stop,ls}}$ were selected by evaluating the models run on adaptively adjusted $10 \times 10$ grids on an evaluation data set with $1000$ individuals. In all setups the grid ran through an equally spaced sequence from $30$ to $300$ in both directions -- for $m_\text{stop,l}$ and $m_\text{stop,ls}$. When analysing these results we noticed that the steps between the vlues on the grid were to wide for setup S2 and S3. The grid was consequently adapted such that $m_\text{stop,l}$ was chosen from an equally spaced sequence from $95$ to $140$, while the grid for $m_\text{stop,ls}$ ran from $130$ to $220$. The optimal stopping iterations were chosen such that the joint log likelihood was maximal on the patients left out of the fitting process (\textit{predictive risk}). For an overview over the resulting stopping iterations see Table~\ref{m_stop:tab}.\\ 

\begin{table}
\begin{center}
\begin{tabular}{|rrrr|}
  \hline
 & S1 & S2 & S3 \\ 
  \hline
$m_{\text{stop,l}}$ & $160.2$ & $187.8$ & $187.2$ \\  
   $m_{\text{stop,ls}}$ & $154.8$ & $48.6$ & $30.6$ \\ 
   \hline
\end{tabular}
\caption{\label{m_stop:tab} Average number of stopping iterations for all three simulation setups listed separately for the two different sub-predictors. }
\end{center}
\end{table}

\section{Cystic Fibrosis data}\label{sec:data}

Cystic fibrosis is the most common life-limiting inherited disease among Caucasian populations, with most patients dying prematurely from respiratory failure. Children with cystic fibrosis in the UK are usually diagnosed in the first year of life, and subsequently require intensive support from family and healthcare services \citep{dtr2013}. Though cystic fibrosis affects other organs such as the pancreas and liver, it is the lung function that is of most concern, with gradual decline in function for most patients leading to the necessity of lung transplantation for many. Lung function decline in cystic fibrosis is accelerated if patients become infected with a range of infections \citep{qvist}. However, the direction of causation is unclear. It may also be the case that accelerated lung function decline predisposes patients to risk of lung infection. There is thus utility in analyzing lung function decline, and time to lung infection in a joint model in order to gain more insight in the structure of the process.

We analyzed part of a dataset from the Danish cystic fibrosis registry. To achieve comparability between patients we chose to use one observation per patient per year. Lung function, measured in forced expiratory volume in one second (FEV1), is the longitudinal outcome of our joint model. The event in the survival part of the model is the onset of the pulmonary infection {\textit{Pseudomonas aeruginosa}} (PA). After selecting the patients that have at least two observations before the infection, the data set contained a total of $5425$ observations of $417$ patients of which $48$ were infected with PA in the course of the study. The mean waiting time until infection was $19.86$, hence the mean age of infection was $24.86$ years, since patients are included at the age of five into the study. The covariates for the longitudinal predictor were height and weight of the patient as well as two binary indicators: one states if the patient had one of three different additional lung infections and a second indicates if the patient had diabetes. The covariates possibly having an impact on the shared part of the model were time (i.e. age of the individuals), pancreatic insufficiency, sex, and age at which cystic fibrosis was diagnosed. Covariates were standardized to have mean zero and standard deviation one, except for age, which was normed on an interval from zero to one.

\begin{figure}[t]\centering
\includegraphics[width=\textwidth]{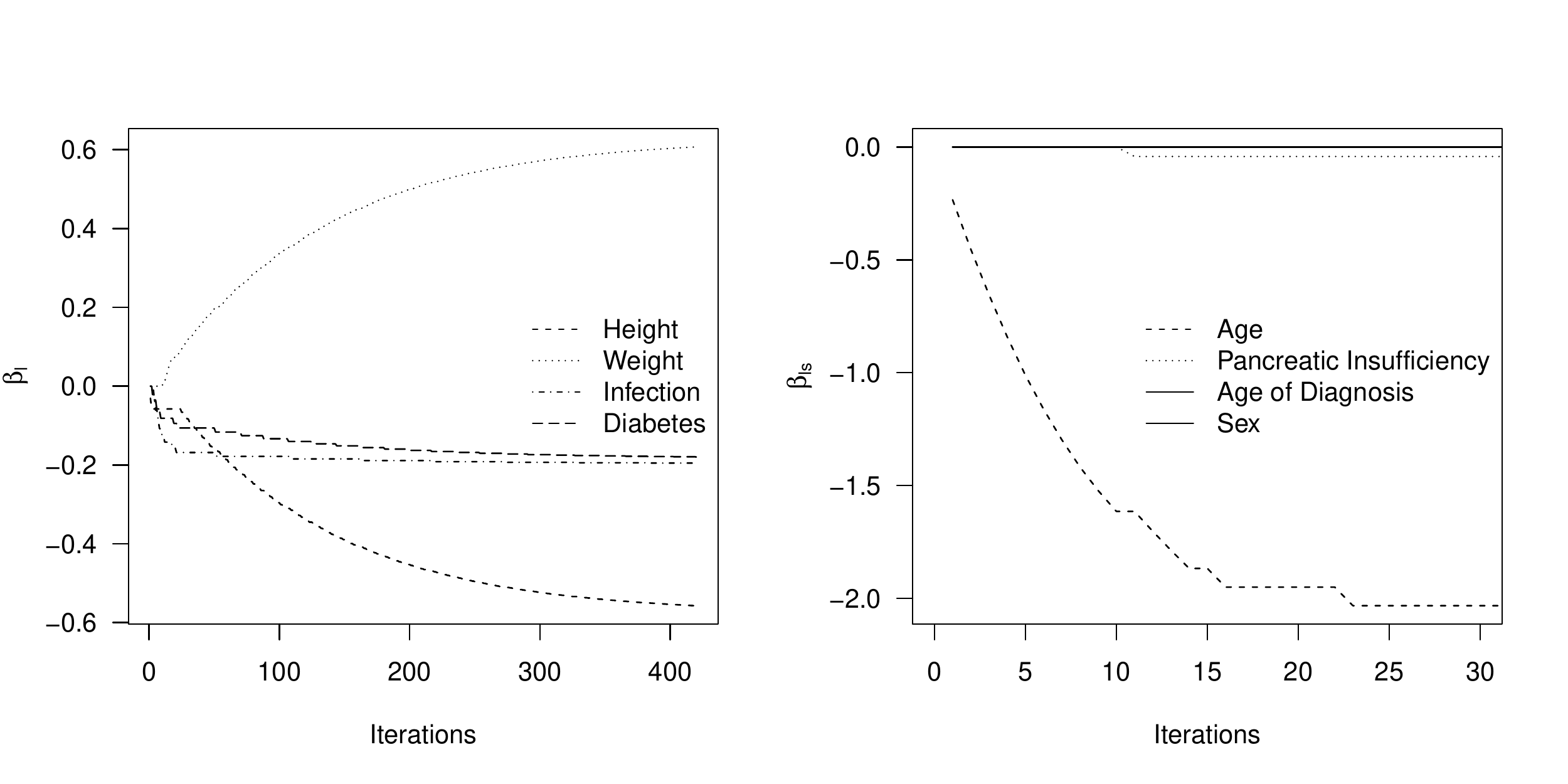}
\caption{\label{coeff_path} Coefficient paths for the fixed effects in the cystic fibrosis model. The graph on the left shows the coefficient paths for the longitudinal sub-predictor (displaying the influence on to the patients lung function), the graph on the right the coefficient paths of the shared sub-predictor (incorporated both in the longitudinal model for the lung function and the survival model for the Pseudomonas infection).}
\end{figure}

We then ran our boosting algorithm on the data set in order to simultaneously estimate and select the most influential variables for the two sub-predictors while optimizing the association parameter $\alpha$. As recommended, we used a fixed step length of 0.1 for both predictors and optimized $m_{\text{stop}}$ instead. The stopping iterations ($m_{\text{stop,l}} = 420$ and $m_{\text{stop,ls}}=30$)  were chosen based on tenfold cross validation via a two-dimensional $15\times 15$ grid, sequencing in equidistant steps of $30$ from $30$ to $450$. The resulting coefficient paths for both sub-predictors are displayed in Figure~\ref{coeff_path}.

The selected variables chosen to be informative for the joint model were \textit{height}, \textit{weight}, \textit{(additional) infection}, \textit{age}, \textit{pancreatic insufficiency} and \textit{diabetes}. \textit{Age of diagnosis} and \textit{sex} did not show to have an impact and were not selected. Being diabetic results to have a negative impact on the lung function, the same holds for having an additional lung infection or pancreatic insufficiency. The negative impact of \textit{age} on the lung function was to be expected since lung function typically declines for patients getting older. The resulting negative impact of \textit{height} and the positive impact of \textit{weight} indicate, that following our joint model, underweight has a negative influence on the lung function.

The association parameter $\alpha$ is negative with a value of $-0.520$. The risk of being infected is hence implicitly -- i.e. via the shared predictor -- associated negatively with the value of the lung function. Lower lung function thus leads to a higher risk of being infected with pseudomonas emphasizing the need to model both outcomes simultaneously in the framework of joint models. 
\section{Discussion and Outlook}

Joint modelling is a powerful tool for analysing longitudinal and survival data, as documented by \cite{riz_book}. It is increasingly applied in biomedical research, yet it was unfeasible for high-dimensional data and  there have been no suggestions for automated variable selection until now. Statistical boosting on the other hand is a modern machine-learning method for variable selection and shrinkage in statistical models for potentially high-dimensional data. We present a possibility to merge these two state of the art statistical fields. Our suggested method leads to an automated selection of informative variables and can deal with high-dimensional data and, probably even more importantly, opens up a whole new range of possibilities in the field of joint modelling.

Boosting algorithms utilizing the underlying base-learners are very flexible when it comes to incorporating different types of effects in the final models. Due to the modular nature of boosting \citep{buhlmann2014discussion}, almost any base-learner can be included in any algorithm -- although the regression setting defined by the underlying loss function might be extremely different. In our context, the proposed boosting algorithm for joint models could facilitate the construction of joint models including not only linear effects and splines, but for example, also spatial and regional effects.

A second area in which the suggested model calls for extension is variable selection itself. While informative variables are selected reliably in most cases, also non informative variables are included in too many iterations. With this being a structural problem of boosting algorithms \citep{BuehlmannYu2007} in low-dimensional settings, the solution could be stability selection \citep{Meinshausen:2008, shah2013variable}, as shown to be helpful in other cases \citep{hofner2015stabsel, mayr2016stabsel}. The difference in the proportion of falsely selected variables between longitudinal and shared sub-predictor however, could be a joint modelling inherent problem and should be subject of future analysis. Further research is also warranted on theoretical insights, as it remains unclear if the existing findings on consistency of boosting algorithms \citep{zhang2005boosting,Buehlmann_2006} hold also for the adapted version for boosting joint models.
 
We plan to extend the model by incorporating a predictor $\eta_s$, i.e. including covariates which only have an impact on the survival time and are independent of the longitudinal structure. Once this is incorporated, we can make even better use of the features of boosting and implement variable selection and allocation between the predictors. The latter is especially useful if no prior knowledge on the structure of association exists, because covariates can be suggested for all three sub-predictors and the algorithm automatically assigns the informative covariates to the appropriate sub-predictor.

In conclusion we propose a new  statistical inference scheme for joint models that also provides a starting point for a much wider framework of boosting joint models, covering a great range of potential models and types of predictor effects.

\subsection*{Acknowledgements}
The work on this article was supported by the German Research Foundation
(DFG), grant SCHM 2966/1-2, grant KN 922/4-2 and the Interdisciplinary Center for Clinical
Research (IZKF) of the Friedrich-Alexander-University Erlangen-N\"urnberg
(Projects J49 and J61). We furthermore want to thank the Editor and the reviewers, especially the one who saved us from falling into the trap of oversimplifying the gradient in our first draft of the algorithm.

\appendix
\section{Gradient of the shared predictor}\label{appendix:calc}

In the following we will sketch out the calculus for the gradient of the likelihood of the shared sub-predictor, indices for individuals/obsevations (i.e. $ij$ will be omitted. Note that the derivation of the first part in equation (\ref{eq:loss}) with respect to $\eta_{\text{ls}}$ is the same as in the purely longitudinal predictor $\frac{1}{\sigma^2}\left(y-\eta_{\text{l}}-\eta_{\text{ls}}\right)$. The additional part in the likelihood of the shared sub-predictor is simply the one of a pure survival model
\begin{eqnarray*}
f(T, \delta|\eta_{\text{ls}},\lambda, \alpha)&=&\left[\lambda\exp(\alpha\eta_{\text{ls}})\right]^{\delta}\exp\left[-\lambda\int_0^{T}\exp(\alpha\eta_{\text{ls}})\ \text{d}t\right].\nonumber
\end{eqnarray*}
The formulation of the log likelihood is thus: 
\begin{eqnarray*}
\log\left(f(T, \delta|\eta_{\text{ls}}, \lambda, \alpha)\right)&=&\underset{(l1)}{\underbrace{\delta\log(\lambda)+ \delta\alpha\eta_{\text{ls}}}}-\underset{(l2)}{\underbrace{\lambda\int_0^{T}\exp(\alpha(\eta_{\text{ls}}))\text{d}t}}.
\end{eqnarray*}
This again splits into two part ($l1$) and ($l2$). The derivation of the first part ($l1$) is straight forward and reduces to $\delta\alpha$. The part of the likelihood less straight forward to differentiate is the part including the integral ($l2$)
\begin{eqnarray*}
\frac{d}{d\eta_{\text{ls}}}\lambda\int_{0}^{T}\exp\left(\alpha\eta_{\text{ls}} \right)\text{d}t.
\end{eqnarray*}
Note that $\eta_{\text{ls}}$ is a function of time such that standard derivation cannot be used. To avoid confusion with the upper integral bound $T$ and the integration variable $\text{d}t$ above, we suppress the argument time in the following. Applying the rules for functional derivatives, we obtain
\begin{eqnarray*}
\frac{d}{d\eta_{\text{ls}}}\lambda{\int_{0}^{T}\exp\left(\alpha\eta_{\text{ls}} \right)\text{d}t}=\lambda\int_{0}^{T}\alpha\exp\left(\alpha\eta_{\text{ls}} \right)\text{d}t.
\end{eqnarray*}
In the case of interest in this paper, where only the second part of the predictor is time-dependent, i.e.,  $\eta_{\text{ls}} = \eta_{\text{ls}-t} + (\beta_t + \gamma_1) t$, the functional derivative of the function $\eta_{\text{ls}}$  has the following form:
\begin{eqnarray*}
\int_{0}^{T}\exp\left(\alpha(\eta_{\text{ls}-t} + (\beta_t  + \gamma_1)t) \right)\text{d}t&=&\left[\frac{\alpha}{\alpha(\gamma_1 + \beta_t)}\exp(\alpha\eta_{\text{ls}-t} +\alpha(\beta_t  + \gamma_1)t)\right]_0^{T}\nonumber\\
&=&\frac{\exp(\alpha\eta_{\text{ls}-t} + \alpha(\beta_t  + \gamma_1)T)-\exp(\alpha\eta_{\text{ls}-t})}{\gamma_1 + \beta_t}.\nonumber\\
\end{eqnarray*}
The whole derivation of the likelihood for the shared subpredictor hence is
\begin{eqnarray*}
\frac{1}{\sigma^2}\left(y-\eta_{\text{l}}-\eta_{\text{ls}}\right)+\delta\alpha-\lambda\frac{\exp(\alpha\eta_{\text{ls}})-\exp(\alpha\eta_{\text{ls}-t})}{\gamma_1 + \beta_t}.
\end{eqnarray*}
\section{Simulation}\label{appendix:sim}

\begin{itemize}	
\item Choose a number of patients $N$ and a number of observations $n_i$ per patient $i, i=1,\ldots,n$
\item  Simulate the time points:
 \begin{itemize}
	\item draw $n=N\cdot n_i$ days in $1,\ldots, 365$
	\item generate the sequence $\boldsymbol{t}_i$ of length $n_i$ years per patient, by taking the above simulated values as days of the year
	\item to make calculations easier, divide everything by 365
	\item example: time points for patient $i$ given that $n_i =5 $: $\boldsymbol{t}=(.2,1.6,2.4,3.1,4.8)$ means the patient was seen in the first year at day $0.2 \cdot 365$, in year two at day $0.6 \cdot 365$ and so on
 \end{itemize}
\item Generate the longitudinal predictor $\boldsymbol{\eta}_{\text{l}}$
\begin{itemize}
	\item generate fixed effects by choosing appropriate values for the covariate matrix $\boldsymbol{X}_{\text{l}}$
	\item the covariate matrix can also include the time vector
	\item  and deciding for values for the parameter vector $\boldsymbol{\beta}_{\text{l}}$
	\item calculate $\boldsymbol{\eta}_{\text{l}} = \boldsymbol{X}_{\text{l}}\boldsymbol{\beta}_{\text{l}}$
  \end{itemize}   	
\item Generate the shared predictor $\boldsymbol{\eta}_{\text{ls}}$ outcomes based on the time points (just as in a usual longitudinal setup):
 \begin{itemize}
	\item generate random intercept $\gamma_0$ and random slope $\gamma_1$ by drawing form a standard Gaussian distribution
  \item generate fixed effects by choosing appropriate values for the covariate matrix $\boldsymbol{X}_{\text{ls}}$ (note that the values are not allowed to be time varying)
	\item  and deciding for values for the parameter vector $\boldsymbol{\beta}_{\text{ls}}$
	\item calculate $\boldsymbol{\eta}_{\text{ls}} = \boldsymbol{X}_{\text{ls}}\boldsymbol{\beta}_{\text{ls}}+ \boldsymbol{\gamma_0} + \boldsymbol{\gamma_1} \boldsymbol{t}$
  \end{itemize}   	
\item draw $n$ values of $\boldsymbol{y}$ from $N\left(\boldsymbol{\eta}_{\text{ls}}(t) + \boldsymbol{\eta}_{\text{l}}(t) ,\sigma^2\right)$
\item Simulate event times (based on the above simulated times and random effects):
 \begin{itemize}	
  \item choose a baseline hazard $\lambda_0$ (for simplicity reasons chosen to be constant) and an association parameter $\alpha$
	\item calculate the probability of an event happening up to each time point $t_{ij}$ with the formula $F_{t_{ij}} =1- \exp\left(-\lambda_0 \frac{\exp\left(\alpha\boldsymbol{\eta}_{\text{ls}}(t)\right)-
\exp(\alpha\boldsymbol{\eta}_{\text{ls}-t})}{\alpha(\beta_t+\gamma_1)}\right)$
	\item draw $n$ uniformly distributed variables $u_{ij}$
	\item if $u_{ij} < F_{t_{ij}}$ consider an event having happened before $t_{ij}$ 
	\item define time of event $s_i$ as proportional to the difference of $u_{ij}$ and $F_{t_{ij}}$ between $t_{ij}$ and $t_{ij-1}$
	\item for every individual $i$ only the first $s_i$ is being considered
	\item define the censoring vector $\boldsymbol{\delta}$ of length $N$ with $\delta_i=0$ if $u_{ij} < F_{t_{ij}}$ for any $j = 1,\ldots, n_i$ and $\delta_i=1$ otherwise (this leads to an average censoring rate of $83.6\%$).
\end{itemize}
 \item for every individual $i$ delete all $y(t_{ij})$ with $t_{ij}\geq s_i$ from the list of observations
\end{itemize}

The probabilities for the events are taken from the connection between the hazard rate and the survival function.
\begin{eqnarray*}
F(t)&=&1 - S(t)\nonumber\\
&=&1 - \exp\left(-\int_{0}^t\lambda(t)\ \text{d}t\right)\nonumber\\
&=& 1- \exp\left(-\lambda_0 \int_0^t \exp\left(\alpha\eta_{\text{ls}}(t)\right)\ \text{d}t\right)\nonumber\\
&=& 1- \exp\left(-\lambda_0 \frac{\exp\left(\alpha\eta_{\text{ls}}(t)\right)-
\exp(\alpha\eta_{\text{ls}-t})}{\alpha(\beta_t+\gamma_1)}\right)\nonumber\\
\end{eqnarray*}

\bibliographystyle{chicago}
\bibliography{library}

\end{document}